%% file: main.tex
\documentclass{article}

\PassOptionsToPackage{numbers, compress}{natbib}
\usepackage[preprint]{neurips_2022}

\usepackage[utf8]{inputenc} 
\usepackage[T1]{fontenc}    
\usepackage{hyperref}       
\usepackage{url}            
\usepackage{booktabs} 

\usepackage{hyperref}

\usepackage{algorithm}
\usepackage{algorithmic}
\usepackage{amsmath}
\usepackage{amssymb}
\usepackage{mathtools}
\usepackage{multicol}
\usepackage{multirow}
\usepackage{pifont}
\newcommand{\cmark}{\ding{51}}%
\newcommand{\xmark}{\ding{55}}%
\usepackage{amsthm}
\usepackage{xcolor}
\usepackage{bbm}
\usepackage{caption}

\usepackage{subcaption}

\usepackage[bottom]{footmisc}

\usepackage[capitalize,noabbrev]{cleveref}

\theoremstyle{plain}

\theoremstyle{definition}

\theoremstyle{remark}

\DeclareMathOperator*{\argmax}{arg\,max}

\makeatletter
\def\algbackskip{\hskip-\ALG@thistlm}
\makeatother

\newcommand{\Dtest}{\mathcal{D}^{\text{test}}}
\newcommand{\Dtrain}{\mathcal{D}^{\text{train}}}
\newcommand{\Dtotal}{\mathcal{D}^{\text{total}}}

\usepackage[textsize=tiny]{todonotes}

\title{Learning to Split for Automatic Bias Detection}
\author{%
  Yujia Bao \\
  CSAIL MIT \\
  \texttt{yujia@csail.mit.edu} \\
  \And
  Regina Barzilay\\
  CSAIL MIT \\
  \texttt{regina@csail.mit.edu}
}

\begin{document}

\maketitle

\begin{abstract}

Classifiers are biased when trained on biased datasets. As a remedy, we propose Learning to Split (\texttt{ls}), an algorithm for automatic bias detection. Given a dataset with input-label pairs, \texttt{ls} learns to split this dataset so that predictors trained on the training split \emph{cannot generalize} to the testing split. This performance gap suggests that the testing split is under-represented in the dataset, which is a signal of potential bias. Identifying non-generalizable splits is challenging since we have no annotations about the bias.
In this work, we show that the prediction correctness of each example in the testing split can be used as a source of weak supervision: generalization performance will drop if we move examples that are predicted correctly away from the testing split, leaving only those that are mispredicted.
\texttt{ls} is task-agnostic and can be applied to any supervised learning problem, ranging from natural language understanding and image classification to molecular property prediction.
Empirical results show that \texttt{ls} is able to generate astonishingly challenging splits that correlate with human-identified biases. Moreover, we demonstrate that combining robust learning algorithms (such as group DRO) with splits identified by \texttt{ls} enables automatic de-biasing. Compared to previous state-of-the-art, we substantially improve the worst-group performance (23.4\% on average) when the source of biases is unknown during training and validation.
Our code and data is available at \url{https://github.com/yujiabao/ls}.
\end{abstract}

\input{sections/introduction}
\input{sections/related}
\input{sections/method}
\input{sections/experiments}
\input{sections/conclusion}



\bibliography{main}

\input{appendix/experiments}
\input{appendix/details}
\input{appendix/noisy}
\input{appendix/full}

\end{document}

%% file: sections/introduction.tex
\section{Introduction}\label{sec:introduction}

Recent work has shown promising results on de-biasing when the sources of bias (e.g., gender, race) are known a priori~\cite{ren2018learning, sagawa2019distributionally, clark2019don, he2019unlearn, mahabadi2020end, kaneko2021debiasing}. 
However, in the general case, identifying bias in an arbitrary dataset may be challenging even for domain experts:
it requires expert knowledge of the task and details of the annotation protocols~\cite{zellers2019hellaswag, sakaguchi2020winogrande}.
In this work, we study \emph{automatic bias detection}: given a dataset with only input-label pairs, our goal is to detect biases that may hinder predictors' generalization performance.

\begin{figure}[t]
\centering
\includegraphics[width=\linewidth]{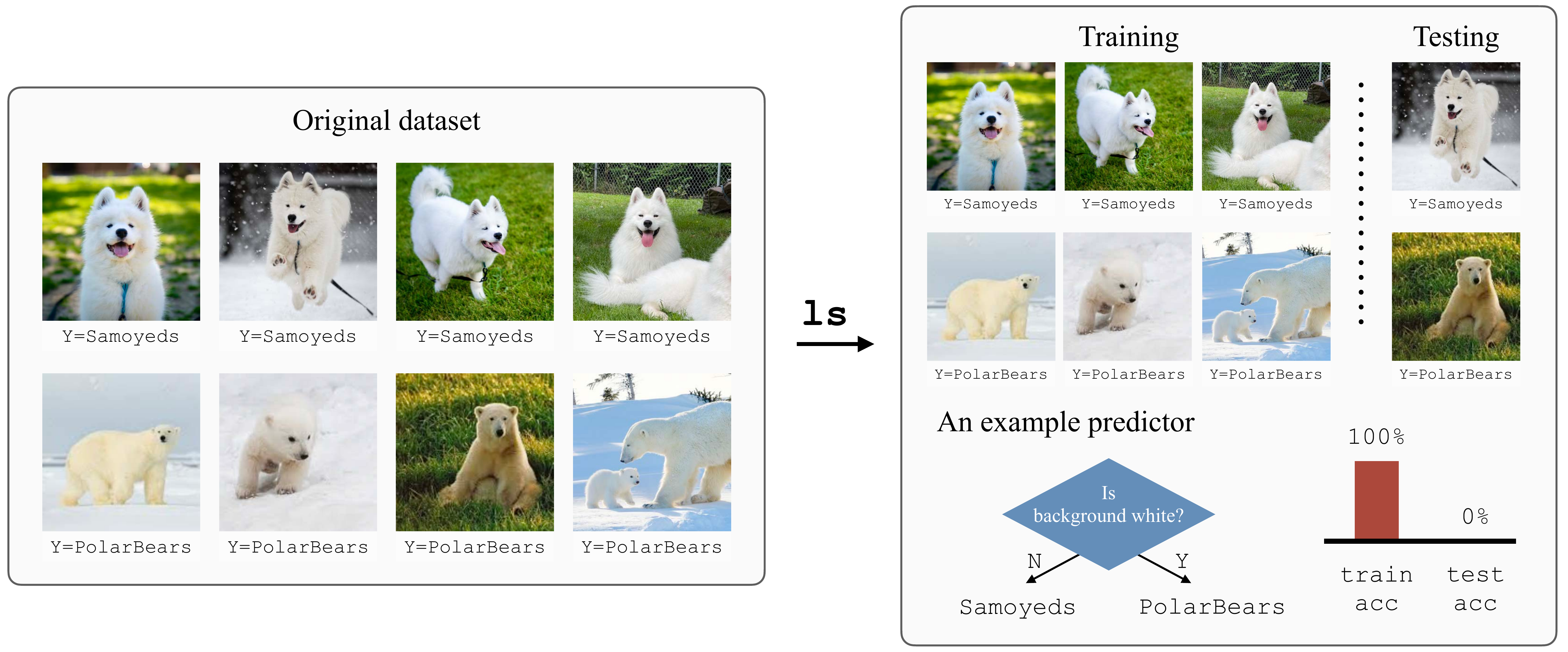}
\caption{Consider the task of classifying samoyed images vs. polar bear images. Given the set of image-label pairs, our algorithm \texttt{ls} \emph{learns} to \emph{split} the data so that predictors trained on the training split \emph{cannot generalize} to the testing split. The learned splits help us identify the hidden biases. For example, while predictors can achieve perfect performance on the training split by using the spurious heuristic: polar bears live in snowy habitats, they fail to generalize to the under-represented group (polar bears that appear on grass).}
\label{fig:intro}
\end{figure}

We propose Learning to Split (\texttt{ls}), an algorithm that simulates generalization failure directly from the set of input-label pairs. Specifically, \texttt{ls} learns to split the dataset so that predictors trained on the training split \emph{cannot generalize} to the testing split  (Figure~\ref{fig:intro}). 
This performance gap indicates that the testing split is under-represented among the set of annotations, which is a signal of potential bias.\looseness=-1

The challenge in this seemingly simple formulation lies in the existence of many trivial splits. For example, poor testing performance can result from a training split that is much smaller than the testing split (Figure~\ref{fig:intro_bad_splits}a). Classifiers will also fail if the training split contains all positive examples, leaving the testing split with only negative examples (Figure~\ref{fig:intro_bad_splits}b). The poor generalization of these trivial solutions arise from the lack of training data and label imbalance, and they do not reveal the hidden biases. To ensure that the learned splits are meaningful, we impose two regularity constraints on the splits. First, the size of the training split must be comparable to the size of the testing split. Second, the marginal distribution of the labels should be the similar across the splits.

Our algorithm \texttt{ls} consists of two components, \emph{Splitter} and \emph{Predictor}.
At each iteration, the Splitter first assigns each input-label pair to either the training split or the testing split.
The Predictor then takes the training split and learns how to predict the label from the input. Its prediction performance on the testing split is used to guide the Splitter towards a more challenging split (under the regularity constraints) for the next iteration. Specifically, while we do not have any explicit annotations for creating non-generalizable splits, we show that the prediction correctness of each testing example can serve as a source of weak supervision: generalization performance will decrease if we move examples that are predicted correctly away from the testing split, leaving only those predicted incorrectly.\looseness=-1

\texttt{ls} is task-agnostic and can be applied to any supervised learning problem, ranging from natural language understanding (Beer Reviews, MNLI) and image classification (Waterbirds, CelebA) to molecular property prediction (Tox21).
Given the set of input-label pairs, \texttt{ls} consistently identifies splits across which predictors cannot generalize. For example in MNLI, the generalization performance drops from 79.4\% (split by random) to 27.8\% 
(split by \texttt{ls}) for a standard BERT-based predictor.
Further analysis reveals that our learned splits coincide with human-identified biases.
Finally, we demonstrate that combining group distributionally robust optimization (DRO) with splits identified by \texttt{ls} enables automatic de-biasing. Compared with previous state-of-the-art, we substantially improves the worst-group performance (23.4\% on average) when the sources of bias are completely unknown during training and validation.

%% file: sections/related.tex
\section{Related work}\label{sec:related}

\paragraph{De-biasing algorithms}
Modern datasets are often coupled with unwanted biases~\cite{buolamwini2018gender, schuster2019towards, mccoy-etal-2019-right, yang2019analyzing}. If the biases have already been identified, we can use this prior knowledge to regulate their negative impact~\cite{kusner2017counterfactual, hu2018does, oren2019distributionally, belinkov2019don, stacey-etal-2020-avoiding, clark2019don, he2019unlearn, mahabadi2020end, Sagawa*2020Distributionally, singh2021anatomizing}.
The challenge arises when the source of biases is unknown~\citep{li-2021-discover}. Recent work has shown that the mistakes of a standard ERM predictor on its \emph{training data} are informative of the biases~\citep{yujia2021predict, sanh2021learning, nam2020learning, utama2020towards, liu2021just, lahoti2020fairness, liu2021heterogeneous}. They deliver robustness by boosting from the mistakes. \citep{creager2021environment, sohoni2020no, ahmed2020systematic, matsuura2020domain} further analyze the predictor's hidden activations to identify under-represented groups.  However, many other factors (such as the initialization, the representation power, the amount of annotations, etc) can contribute to the predictors' training mistakes. For example, predictors that lack representation power may simply under-fit the training data.

In this work, instead of looking at the training statistics of the predictor, we focus on its \emph{generalization gap} from the training split to the testing split. This effectively balances those unwanted factors. 
Going back to the previous example, if the training and test splits share the same distribution, the generalization gap will be small even if the predictors are underfitted.
The gap will increase only when the training and testing splits have \emph{different} prediction characteristics. Furthermore, instead of using a fixed predictor, we iteratively refine the predictor during training so that it faithfully measures the generalization gap given the current Splitter.

\begin{figure*}[t]
    \centering
    \includegraphics[width=\linewidth]{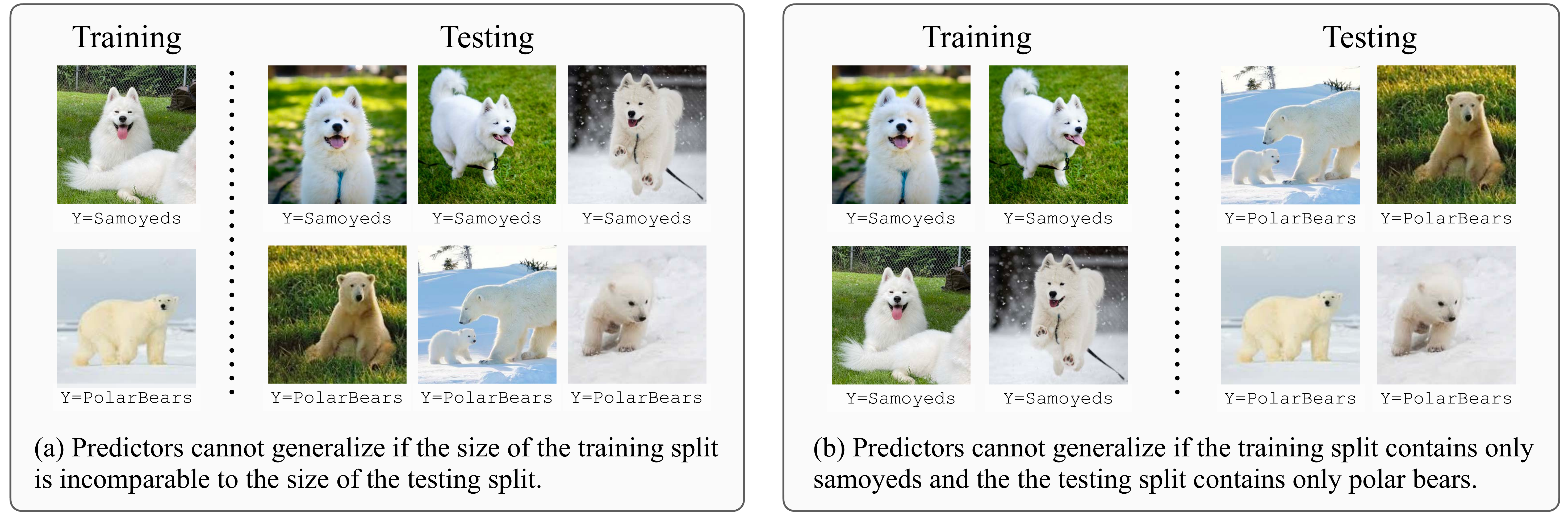}
    \caption{Splits that are difficult to generalize do not necessarily reveal hidden biases. (a) Predictors cannot generalize if the amount of annotations is insufficient. (b) Predictors fail to generalize when the labels are unbalanced in training and testing.
    \texttt{ls} poses two regularity constraints to avoid such degenerative solutions: the training split and testing split should have comparable sizes; the marginal distribution of the label should be similar across the splits.}
    \label{fig:intro_bad_splits}
    \vspace{-3mm}
\end{figure*}

\paragraph{Heuristics for data splitting}
Data splitting strategy directly impacts the difficulty of the underlying generalization task.
Therefore, in domains where out-of-distribution generalization is crucial for performance, various heuristics are used to find challenging splits~\cite{sheridan2013time, yang2019analyzing,bandi2018detection, yala2021multi,taylor2019rxrx1, koh2021wilds}. Examples include scaffold split in molecules and batch split for cells.
Unlike these methods, which rely on human-specified heuristics, our algorithm \texttt{ls} learns how to split directly from the dataset alone and can therefore be applied to scenarios where human knowledge is unavailable or incomplete.

\paragraph{Meta learning}
Learning to Split (\texttt{ls}) naturally involves a bi-level optimization problem~\citep{franceschi2018bilevel, sinha2017review, hospedales2020meta}. Past work has successfully applied meta learning to learn model initializations~\citep{finn2017model}, optimizers~\citep{andrychowicz2016learning}, metric spaces~\citep{vinyals2016matching, snell2017prototypical}, network architectures~\citep{liu2018darts}, instance weights~\citep{ren2018learning, jiang2018mentornet, shu2019meta}, teaching policies~\citep{fan2018learning}. In these methods, the inner-loop and outer-loop models cooperate with each other. In this work, our outer-loop Splitter  plays an adversarial game~\citep{goodfellow2014generative} against the inner-loop Predictor. We learn how to split the data so that predictors will \emph{fail} to generalize.

%% file: sections/method.tex
\section{Learning to Split}\label{sec:method}
\subsection{Motivation}
Given a dataset $\mathcal{D}^{\text{total}}$ with input-label pairs $\{(x, y)\}$, our goal is to split this dataset into two subsets, $\mathcal{D}^{\text{train}}$ and $\mathcal{D}^{\text{test}}$, such that predictors learned on the training split $\mathcal{D}^{\text{train}}$ cannot generalize to the testing split $\mathcal{D}^{\text{test}}$.\footnote{To prevent over-fitting, we held-out 1/3 of the training split for early-stopping when training the Predictor.}

\emph{\textbf{Why do we have to discover such splits?}} Before deploying our trained models, it is crucial to understand the extent to which these models can even generalize within the given dataset. The standard cross-validation approach attempts to measure generalization by randomly splitting the dataset~\citep{stone1974cross, allen1974relationship}. However, this measure only reflects the \emph{average} performance under the same data distribution $\mathbb{P}_{\Dtotal}(x,y)$.
There is no guarantee of performance if our data distribution changes at test time (e.g. increasing the proportion of the minority group).
For example, consider the task of classifying samoyeds vs. polar bears (Figure~\ref{fig:intro}). Models can achieve good average performance by using spurious heuristics such as ``polar bears live in snowy habitats'' and ``samoyeds play on grass''.
Finding splits across which the models cannot generalize helps us identify underrepresented groups (polar bears that appear on grass).

\emph{\textbf{How to discover such splits?}} Our algorithm \texttt{ls} has two components, a \emph{Splitter} that decides how to split the dataset and a \emph{Predictor} that estimates the generalization gap from the training split to the testing split. At each iteration, the splitter uses the feedback from the predictor to update its splitting decision.  One can view this splitting decision as a latent variable that represents the prediction characteristic of each input.
To avoid degenerate solutions, we require the Splitter to satisfy two regularity constraints: the size of the training split should be comparable to the size of the testing split (Figure~\ref{fig:intro_bad_splits}a); and the marginal distribution of the label should be similar across the splits (Figure~\ref{fig:intro_bad_splits}b).

\begin{algorithm}[t]
    \caption{Learning to Split (\texttt{ls})}
    \label{alg:ls}    
    {\bfseries Input:} dataset $\Dtotal$\\
    {\bfseries Output:} data splits $\Dtrain$, $\Dtest$
    \begin{algorithmic}[1]
        \STATE Initialize \emph{Splitter} as random splitting
        \REPEAT
        \STATE Apply \emph{Splitter} to split $\Dtotal$ into $\Dtrain, \Dtest$. 
        \STATE Initialize \emph{Predictor} and train \emph{Predictor} on $\Dtrain$ using empirical risk minimization.
        \STATE Evaluate \emph{Predictor} on $\Dtest$ and compute its generalization \texttt{gap}.
        \REPEAT
            \STATE Sample a mini-batch from $\Dtotal$ to compute the regularity constraints $\Omega_1, \Omega_2$  (Eq~\ref{eq:reg}).
            \STATE Sample another mini-batch from $\Dtest$ to compute $\mathcal{L}^{\text{gap}}$  (Eq~\ref{eq:gap}).
            \STATE Update \emph{Splitter} to minimize the overall objective $\mathcal{L}^{\text{total}}$ (Eq~\ref{eq:total}).
        \UNTIL{$\mathcal{L}^{\text{total}}$ stops decreasing}
        \UNTIL{\texttt{gap} stops increasing}
    \end{algorithmic}
\end{algorithm}

\subsection{Splitter and Predictor}
Here we describe the two key components of our algorithm, \emph{Splitter} and \emph{Predictor}, in the context of classification tasks. The algorithm itself generalizes to regression problems as well.

\paragraph{Splitter}
Given a list of input-label pairs $\Dtotal=[(x_1, y_2),\ldots, (x_n, y_n)]$, the Splitter decides how to partition this dataset into a training split $\Dtrain$ and a testing split $\Dtest$. We can view its splitting decisions as a list of latent variables $\mathbf{z}=[z_1,\ldots,z_n]$ where each $z_i\in\{1, 0\}$ indicates whether example $(x_i, y_i)$ is included in the training split or not. In this work, we assume independent selections for simplicity. That is,
the Splitter takes one input-label pair $(x_i, y_i)$ at a time and predicts the probability $\mathbb{P}_{Splitter}(z_i \mid x_i, y_i)$ of allocating this example to the training split. We can factor the joint probability of our splitting decisions as
\[
\mathbb{P}(\mathbf{z} \mid \Dtotal) = \prod_{i=1}^n \mathbb{P}_{Splitter}(z_i \mid x_i, y_i).
\]
We can sample from the Splitter's predictions $\mathbb{P}_{Splitter}(z_i \mid x_i, y_i)$ to obtain the splits $\Dtrain$ and $\Dtest$. Note that while the splitting decisions are independent across different examples, the Splitter receives \emph{global} feedback, dependent on the entire dataset $\Dtotal$, from the Predictor during training.

\paragraph{Predictor}
The Predictor takes an input $x$ and predicts the probability of its label $\mathbb{P}_{Predictor}(y\mid x)$. The goal of this Predictor is to provide feedback for the Splitter so that it can generate more challenging splits at the next iteration.

Specifically, given the Splitter's current splitting decisions, we re-initialize the Predictor and train it to minimize the empirical risk on the training split $\Dtrain$.
This re-initialization step is critical because it ensures that the predictor does not carry over past information (from previous splits) and faithfully represents the current generalization gap.
On the other hand, we note that neural networks can easily remember the training split. To prevent over-fitting, we held-out 1/3 of the training split for early stopping.
After training, we evaluate the generalization performance of the Predictor on the testing split $\Dtest$.

\subsection{Regularity constraints}\label{sec:regularity}
Many factors can impact generalization, but not all of them are of interest. For example, the Predictor may naturally fail to generalize due to the lack of training data or due to label imbalance across the splits (Figure~\ref{fig:intro_bad_splits}). To avoid these trivial solutions, we introduce two regularizers to shape the Splitter's decisions:
\begin{equation}\label{eq:reg}
\begin{aligned}
    \Omega_1 &= \mathrm{D}_{KL}
        (\mathbb{P}(z)\, \|\, \mathrm{Bernoulli}(\delta)
        ),\\
    \Omega_2 &=  \mathrm{D}_{KL}(\mathbb{P}(y\mid z=1)\,\|\, \mathbb{P}(y)) + \mathrm{D}_{KL}(\mathbb{P}(y\mid z=0)\,\|\, \mathbb{P}(y)).
\end{aligned}
\end{equation}

The first term $\Omega_1$ ensures that we have sufficient training examples in $\Dtrain$. Specifically, the marginal distribution $\mathbb{P}(z) = \frac{1}{n} \sum_{i=1}^n \mathbb{P}_{Splitter}(z_i=z \mid x_i, y_i)$ represents what percentages of $\Dtotal$ are split into $\Dtrain$ and $\Dtest$. We penalize the Splitter if it moves too far away from the prior distribution $\mathrm{Bernoulli}(\delta)$.
\citet{centola2018experimental} suggest that minority groups typically make up 25 percent of the population. Therefore, we fix $\delta=0.75$ in all experiments.

The second term $\Omega_2$  aims to reduce label imbalance across the splits. It achieves this goal by pushing the label marginals in the training split  $\mathbb{P}(y \mid z=1)$ and the testing split $\mathbb{P}(y \mid z=0)$ to be close to the original label marginal $\mathbb{P}(y)$ in $\Dtotal$. We can apply Bayes's rule to compute these conditional label marginals directly from the Splitter's decisions $\mathbb{P}_{S.}(z_i\mid x_i, y_i)$:
\[
\mathbb{P}(y \mid z = 1) = \frac{\sum_{i} \mathbbm{1}_{y}(y_i)\, \mathbb{P}_{S.}(z_i = 1\mid x_i, y_i)}{\sum_{i} \mathbb{P}_{S.}(z_i=1\mid x_i, y_i)},
\quad
\mathbb{P}(y \mid z = 0) = \frac{\sum_{i} \mathbbm{1}_{y}(y_i)\, \mathbb{P}_{S.}(z_i = 0\mid x_i, y_i)}{\sum_{i} \mathbb{P}_{S.}(z_i=0\mid x_i, y_i)}.
\]

\subsection{Training strategy}
The only question that remains is how to learn the Splitter.
Our goal is to produce \emph{difficult} and \emph{non-trivial} splits so that the Predictor cannot generalize. However, the challenge is that we don't have any explicit annotations for the splitting decisions.

There are a few options to address this challenge. From the meta learning perspective, we can back-propagate the Predictor's loss on the testing split directly to the Splitter. 
This process is expensive as it involves higher order gradients from the Predictor's training. 
While one can apply episodic-training~\citep{vinyals2016matching} to reduce the computation cost,
the Splitter's decision will be biased by the size of the learning episodes (since the Predictor only operates on the sampled episode). From the reinforcement learning viewpoint, we can cast our objectives, maximizing the generalization gap while maintaining the regularity constraints, into a reward function~\citep{lei2016rationalizing}. However, according to our preliminary experiments,
the learning signal from this scalar reward is too sparse for the Splitter to learn meaningful splits.

In this work, we take a simple yet effective approach to learn the Splitter. Our intuition is that \emph{the Predictor's generalization performance will drop if we move examples that are predicted correctly away from the testing split, leaving only those that are mispredicted}.
In other words, we can view the prediction correctness of the testing example as a direct supervision for the Splitter.\footnote{One may wonder if we can directly use the prediction correctness to create the final split (instead of learning the Splitter). The answer is no, and there are two reasons: 1) It may not satisfy the regularity constraints; 2) It ignores under-represented examples in the original training split that we used to train the Predictor. 
In this work, we parameterize the splitting decisions through a learnable mapping, the Splitter. 
This encourages similar inputs to receive similar splitting decisions, and we can also easily incorporate different constraints into the learning objective (Eq~\ref{eq:total}).
Moreover, by iteratively refining the Predictor based on the updated Splitter, we obtain more challenging splits (Figure~\ref{fig:splitter_learning_curve}).}

Formally, let $\hat{y_i}$ be the Predictor's prediction for input $x_i$: $\hat{y_i} = \argmax_y \mathbb{P}_{Predictor}(y\mid x_i)$.
We minimize the cross entropy loss between the Splitter's decision and the Predictor's prediction correctness over the testing split:
\begin{equation}\label{eq:gap}
    \mathcal{L}^{\text{gap}} = 
    \frac{1}{|\Dtest|} \sum_{(x_i, y_i) \in \Dtest}  \mathcal{L}^{\text{CE}}(\mathbb{P}_{Splitter}(z_i \mid x_i,y_i), \mathbbm{1}_{y_i}(\hat{y_i})).
\end{equation}
Combining with the aforementioned regularity constraints, the overall objective for the Splitter is
\begin{equation}\label{eq:total}
    \mathcal{L}^\text{total} = \mathcal{L}^{\text{gap}} + \Omega_1 + \Omega_2,
\end{equation}

One can explore different weighting schemes for the three loss terms~\citep{pmlr-v80-chen18a}. In this paper, we found that the unweighted summation (Eq~\ref{eq:total}) works well out-of-the-box across all our experiments. Algorithm~\ref{alg:ls} presents the pseudo-code of our algorithm.
At each outer-loop (line 2-11), we start by using the current Splitter to partition $\Dtotal$ into $\Dtrain$ and $\Dtest$. We train the Predictor from scratch on $\Dtrain$ and evaluate its generalization performance on $\Dtest$. For computation efficiency, we sample mini-batches in the inner-loop (line 6-10) and update the Splitter based on Eq~\eqref{eq:total}.

%% file: sections/experiments.tex
\section{Experiments}\label{sec:applications}
In this section, we want to answer two main questions.
\begin{itemize}
    \item Can \texttt{ls} identify splits that are not generalizable? (Section~\ref{sec:exp_ls})
    \item Can we use the splits identified by \texttt{ls} to reduce unknown biases? (Section~\ref{sec:exp_reduce})
\end{itemize}
We conduct experiments over multiple modalities (Section~\ref{sec:exp_data}). Implementation details are deferred to the Appendix. 
Our code is included in the supplemental materials and will be publicly available.

\begin{figure}[t]
    \centering
    \includegraphics[width=\linewidth]{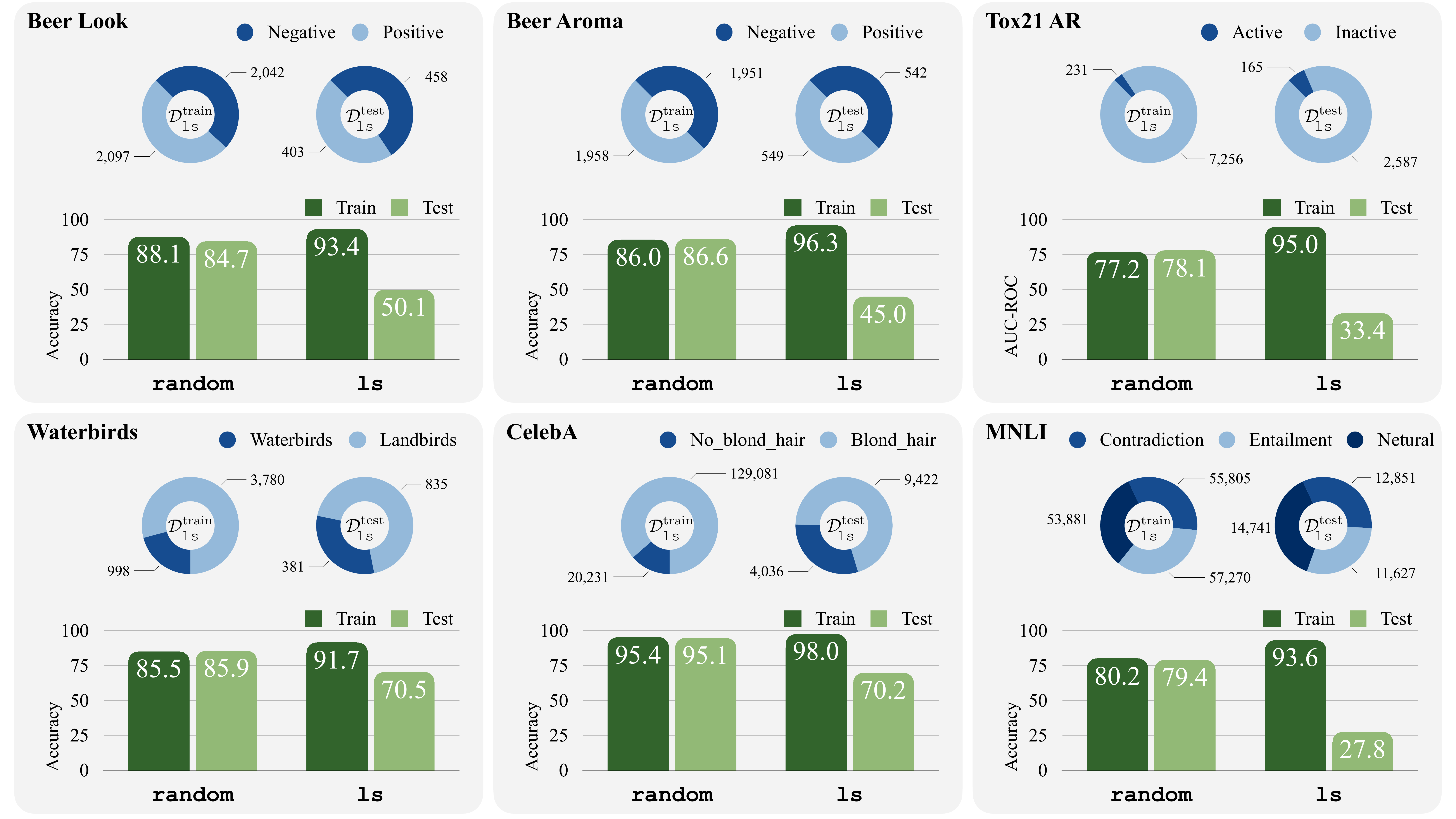}
    \caption{Even while the label distributions remain similar (blue), predictors that generalize on \texttt{random} splits fail to generalize on splits identified by \texttt{ls} (green). For both splits, to prevent over-fitting, we held-out 1/3 of the training split for early-stopping.
    In MNLI (lower right), the generalization gap for a standard BERT-based model is 93.6\%-27.8\%=\textbf{65.8}\%.
    }
    \label{fig:splits}
\end{figure}

\begin{figure}[!t]
    \centering
    \includegraphics[width=\linewidth]{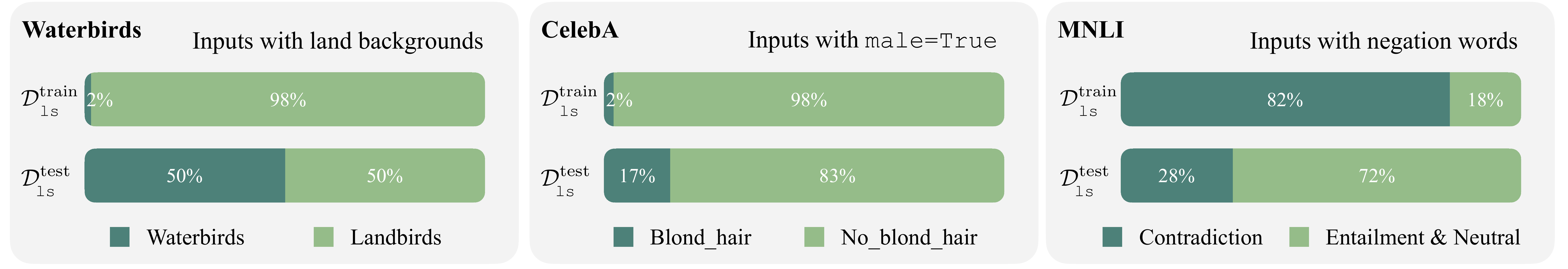}
    \caption{The splits learned by \texttt{ls} correlate with human-identified biases. For example in Waterbirds (left), \texttt{ls} learns to \emph{amplify} the spurious association between landbirds and land backgrounds in the training split $\Dtrain$. As a result, predictors will over-fit the background features and fail to generalize at test time ($\Dtest$) when the spurious correlation is reduced.}
    \label{fig:analysis}
\end{figure}

\begin{figure}[t]
     \centering
     \includegraphics[width=\linewidth]{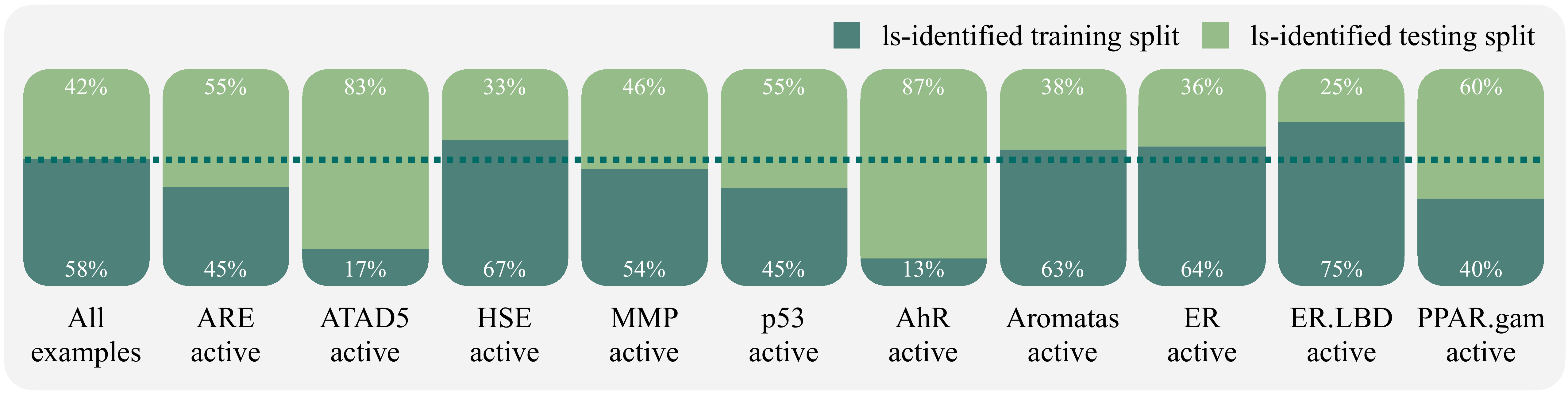}
     \caption{\texttt{ls}-identified splits correlate with certain spurious properties (\texttt{ATAD5}, \texttt{AhR}) even though they are not provided to algorithm. Here we present the train-test assignment of compounds with \texttt{AR=active} given by \texttt{ls}. In the leftmost bar, we look at all examples: 58\% of $\{\texttt{AR=active}\}$ is in the training split and 42\% of $\{\texttt{AR=active}\}$ is in the testing split.
     For each bar on the right, we look at the subset where an unknown property is active.
     For example, 17\% of $\{\texttt{AR=active}, \texttt{ATAD5=active}\}$ is allocated to the training split and 83\% of $\{\texttt{AR=active}, \texttt{ATAD5=active}\}$ is in the testing split.
     }
     \label{fig:tox21}
\end{figure}

\begin{figure}[t]
    \centering
    \includegraphics[width=\linewidth]{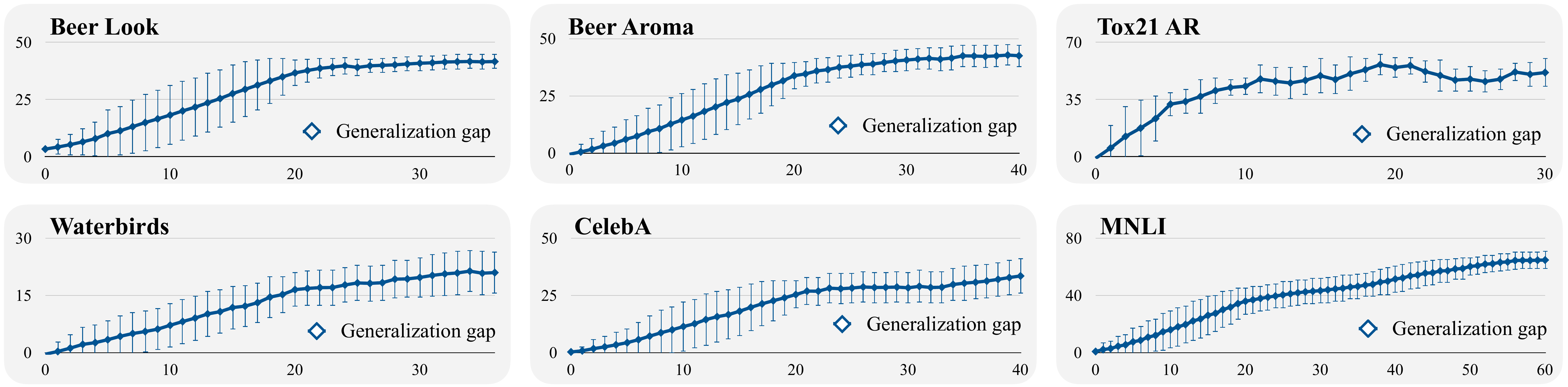}
    \caption{Learning curve of \texttt{ls}.
    X-axis: number of outer-loop iterations. Y-axis: generalization gap from $\Dtrain$ to $\Dtest$. Error bar represents the standard deviation across 5 random seeds.}
    \label{fig:splitter_learning_curve}
    \vspace{-3mm}
\end{figure}

\subsection{Dataset}\label{sec:exp_data}

\noindent\textbf{Beer Reviews}
We use the BeerAdvocate review dataset~\citep{mcauley2012learning} where each input review describes multiple aspects of a beer and is written by a website user. Following previous work~\citep{lei2016rationalizing}, we consider two aspect-level sentiment classification tasks: look and aroma. There are 2,500 positive reviews and 2,500 negative reviews for each task. The average word count per review is 128.5. We apply \texttt{ls} to identify spurious splits for each task.

\noindent\textbf{Tox21} 
Tox21 is a molecular property prediction benchmark with 12,707 chemical compounds~\cite{10.3389/fenvs.2015.00085}.
Each input is annotated with a set of binary properties that represent the outcomes of different toxicological experiments. We consider the property Androgen Receptor ($\texttt{AR}\in \{\texttt{active},\, \texttt{inactive}\}$) as our prediction target. We apply \text{ls} to identify spurious splits over the entire dataset.
\paragraph{Waterbirds}
\citet{sagawa2019distributionally} created this dataset by combining bird images from the Caltech-UCSD Birds-200-2011 (CUB) dataset~\citep{WelinderEtal2010} with backgrounds from the Places dataset~\citep{zhou2014learning}. The task is to predict waterbirds vs. landbirds. The challenge is that waterbirds, by construction, appear more frequently with a water  background. As a result, predictors may utilize this spurious correlation to make their predictions. We combine the official training data and validation data (5994 examples in total) and apply \texttt{ls} to identify spurious splits. 

\noindent\textbf{CelebA}
CelebA is an image classification dataset where each input image (face) is paired with multiple human-annotated attributes~\cite{liu2015deep}.
Following previous work~\citep{sagawa2019distributionally}, we treat the hair color attribute ($y \in \{\texttt{blond}, \texttt{not\_blond}\}$) as our prediction target. The label is spuriously correlated with the gender attribute ($\{\texttt{male}, \texttt{female}\}$). We apply \texttt{ls} to identify spurious splits over the official training data (162,770 examples).

\noindent\textbf{MNLI}
MNLI is a crowd-sourced collection of 433k sentence pairs annotated with textual entailment information~\citep{N18-1101}.
The task is to classify the relationship between a pair of sentences: entailment, neutral or contradiction. 
Previous work has found that contradiction examples often include negation words~\citep{mccoy-etal-2019-right}. We apply \texttt{ls} to identify spurious splits over the training data (206,175 examples) created by \citet{sagawa2019distributionally}.

\subsection{Identifying non-generalizable splits}\label{sec:exp_ls}
Figure~\ref{fig:splits} presents the splits identified by our algorithm \texttt{ls}. Compared to random splitting, \texttt{ls} achieves astonishingly higher generalization gaps across all 6 tasks. Moreover, we observe that the learned splits are not degenerative: the training split $\Dtrain$ and testing split $\Dtest$ share similar label distributions. This confirms the effectiveness of our regularity objectives.

\emph{\textbf{Why are the learned splits so challenging for predictors to generalize across?}}
While \texttt{ls} only has access to the set of input-label pairs, Figure~\ref{fig:analysis} and Figure~\ref{fig:tox21} show that the learned splits are informative of  human-identified biases. For example, in the generated training split of MNLI, inputs with negation words are mostly labeled as contradiction. This encourages predictors to leverage the presence of negation words to make their predictions. These biased predictors cannot generalize to the testing split, where inputs with negation words are mostly labeled as entailment or neutral.

\footnotetext{For fair comparison, all methods share the same hyper-parameter search space and representation backbone (\texttt{resnet-50} for Waterbirds and CelebA, \texttt{bert-base-uncased} for MNLI). See Appendix~\ref{app:details} for details.}

\paragraph{Convergence and time-efficiency} 
\texttt{ls} requires learning a new Predictor for each outer-loop iteration. While this makes \texttt{ls} more time-consuming than training a regular ERM model, this procedure guarantees that the Predictor faithfully measures the generalization gap based on the current Splitter. Figure~\ref{fig:splitter_learning_curve} shows the learning curve of \texttt{ls}. We observe that the generalization gap steadily increases as we refine the Splitter and the learning procedure usually converges within 50 outer-loop iterations.

\subsection{Automatic de-biasing}\label{sec:exp_reduce}
Once \texttt{ls} has identified the spurious splits, we can  apply robust learning algorithms to learn models that generalize across the splits~\citep{sagawa2019distributionally, yujia2021predict}. Here we consider group distributionally robust optimization (group DRO) and study three well-established benchmarks: Waterbirds, CelebA and MNLI.

\input{tables/robust}

\paragraph{Group DRO} Group DRO has shown strong performance when biases are annotated~\citep{sagawa2019distributionally}. For example in CelebA, gender (\texttt{male}, \texttt{female}) constitutes a bias for predicting blond hair. Group DRO uses the gender annotations to partition the training data into four groups: $\{\texttt{blond\_hair}, \texttt{male}\}$, $\{\texttt{blond\_hair}, \texttt{female}\}$, $\{\texttt{no\_blond\_hair}, \texttt{male}\}$, $\{\texttt{no\_blond\_hair}, \texttt{female}\}$. By minimizing the \emph{worst-group} loss during training, it regularizes the impact of the unwanted gender bias. At test time, we report the average accuracy and worst-group accuracy over a held-out test set.

\paragraph{Group DRO with supervised bias predictor}
Recent work consider a more challenging setting where bias annotations are not provided at train time~\citep{levy2020large,nam2020learning, liu2021just,creager2021environment}. 
To achieve robustness, CVaR DRO up-weights examples that have the highest training losses.
LfF and JTT train a separate de-biased predictor by learning from the mistakes of a biased predictor.
EIIL infers the environment information from an ERM predictor and uses group DRO to promote robustness across the latent environments.
However, these methods still access bias annotations on the validation data for model selection.
With thousands of validation examples (1199 for Waterbirds, 19867 for CelebA, 82462 for MNLI), a simple baseline was overlooked by the community: learning a bias predictor over the validation data (where bias annotations are available) and using the predicted bias attributes on the training data to define groups for group DRO.

\paragraph{Group DRO with splits identified by \texttt{ls}}
We consider the general setting where biases are not known during both training and validation. To obtain a robust model, we take the splits identified by \texttt{ls} (Section~\ref{sec:exp_ls}) and use them to define groups for group DRO. For example, we have four groups in CelebA: $\{\texttt{blond\_hair}, z=0\}$, $\{\texttt{blond\_hair}, z=1\}$, $\{\texttt{no\_blond\_hair}, z=0\}$, $\{\texttt{no\_blond\_hair}, z=1\}$. For model selection, we apply the learned Splitter to split the validation data and measure the worst-group accuracy.

\paragraph{Results}
Table~\ref{tab:robust} presents our results on de-biasing. We first see that when the bias annotations are available in the validation data, the missing baseline Group DRO (with supervised bias predictor) outperforms all previous de-biasing methods (4.8\% on average). This result is not surprising given the fact that the bias attribute predictor, trained on the validation data, is able to achieve an accuracy of 94.8\% in Waterbirds (predicting the spurious background), 97.7\% in CelebA (predicting the spurious gender attribute) and 99.9\% in MNLI (predicting the presence of negation words).

When bias annotations are not provided for validation, previous de-biasing methods fail to improve over the ERM baseline, confirming the findings of \citet{liu2021just}. On the other hand, applying group DRO with splits identified by \texttt{ls} consistently achieves the best worst-group accuracy, outperforming previous methods by 23.4\% on average. While we no longer have access to the bias annotations for model selection, the worst-group performance defined by \texttt{ls} can be used as a surrogate (see Appendix~\ref{app:full} for details).

%% file: tables/robust.tex
\begin{table*}[t]
\centering
\caption{Average and worst-group test accuracy for de-biasing.\protect\footnotemark\ Previous work (CVaR DRO, LfF, EIIL, JTT) improve worst-group performances when bias annotations are provided for model selection. However, they underperform the group DRO baseline that was previously overlooked. When bias annotations are not available for validation, the performances of these methods quickly drop to that of ERM. In contrast, applying group DRO with splits identified by \texttt{ls} substantially improves the worst-group performance. $\mathbf{\dagger}$ denotes numbers reported by previous work.}
\small
\label{tab:robust}
\begin{tabular}{lcccccccc}
\toprule\midrule
\multirow{2}{*}{Method} 
& \multicolumn{2}{c}{Bias annotated}    & \multicolumn{2}{c}{Waterbirds} & \multicolumn{2}{c}{CelebA} & \multicolumn{2}{c}{MNLI} \\
\cmidrule(lr{0.5em}){4-5}\cmidrule(lr{0.5em}){6-7}\cmidrule(lr{0.5em}){8-9}
& in train?\hspace{-1mm} & \hspace{-1mm}in val?  & Avg. & Worst & Avg. & Worst & Avg. & Worst \\\midrule
Group DRO{\footnotesize~\citep{sagawa2019distributionally}} & \cmark & \cmark & 93.5\%$^\dagger$ & 91.4\%$^\dagger$ & 92.9\%$^\dagger$ & 88.9\%$^\dagger$ & 81.4\%$^\dagger$ & 77.7\%$^\dagger$ \\ \midrule\midrule
ERM       & \xmark & \cmark & 97.3\%$^\dagger$ & 72.6\%$^\dagger$ & 95.6\%$^\dagger$ & 47.2\%$^\dagger$ & 82.4\%$^\dagger$ & 67.9\%$^\dagger$  \\\midrule
CVaR DRO{\footnotesize~\citep{levy2020large}}  & \xmark & \cmark & 96.0\%$^\dagger$ & 75.9\%$^\dagger$ & 82.5\%$^\dagger$ & 64.4\%$^\dagger$ & 82.0\%$^\dagger$ & 68.0\%$^\dagger$  \\\midrule
LfF{\footnotesize~\citep{nam2020learning}}         & \xmark & \cmark & 91.2\%$^\dagger$ & 78.0\%$^\dagger$ & 85.1\%$^\dagger$ & 77.2\%$^\dagger$ & 80.8\%$^\dagger$   & 70.2\%$^\dagger$     \\\midrule
EIIL{\footnotesize~\citep{creager2021environment}} & \xmark & \cmark & 96.9\%$^\dagger$ & 78.7\%$^\dagger$ & 89.5\%{\phantom{$^\dagger$}} & 77.8\%{\phantom{$^\dagger$}} & 79.4\%{\phantom{$^\dagger$}} & 70.0\%{\phantom{$^\dagger$}} \\\midrule
JTT{\footnotesize~\citep{liu2021just}}             & \xmark & \cmark & 93.3\%$^\dagger$ & 86.7\%$^\dagger$ & 88.0\%$^\dagger$ & 81.1\%$^\dagger$ & 78.6\%$^\dagger$   & 72.6\%$^\dagger$     \\\midrule
Group DRO
& \xmark & \cmark & \\
(with supervised bias predictor)\hspace{-20mm} & & 
& 91.4\%{\phantom{$^\dagger$}} & {\textbf{88.2\%}}{\phantom{$^\dagger$}} & 91.4\%{\phantom{$^\dagger$}} & \textbf{88.9\%}{\phantom{$^\dagger$}} & 79.9\%{\phantom{$^\dagger$}} & \textbf{77.7\%}{\phantom{$^\dagger$}} 
\\\midrule\midrule
ERM       & \xmark  & \xmark & 90.7\%{\phantom{$^\dagger$}} & 64.8\%{\phantom{$^\dagger$}} & 95.8\%{\phantom{$^\dagger$}} & 41.1\%{\phantom{$^\dagger$}} & 81.9\%{\phantom{$^\dagger$}} & 60.4\%{\phantom{$^\dagger$}} \\\midrule
CVaR DRO{\footnotesize~\citep{levy2020large}}       & \xmark  & \xmark &   ---        & 62.0\%$^\dagger$      &  ---      & 36.1\%$^\dagger$ & 81.8\%{\phantom{$^\dagger$}}   & 61.8\%{\phantom{$^\dagger$}}     \\\midrule
LfF{\footnotesize~\citep{nam2020learning}}          & \xmark  & \xmark &   ---        & 44.1\%$^\dagger$      &  ---      & 24.4\%$^\dagger$ & 81.1\%{\phantom{$^\dagger$}}   & 62.2\%{\phantom{$^\dagger$}}     \\\midrule
EIIL{\footnotesize~\citep{creager2021environment}}   & \xmark  & \xmark & 90.8\%{\phantom{$^\dagger$}} & 64.5\%{\phantom{$^\dagger$}} & 95.7\%{\phantom{$^\dagger$}}      & 41.7\%{\phantom{$^\dagger$}}       &    80.3\%{\phantom{$^\dagger$}}      & 64.7\%{\phantom{$^\dagger$}}  \\\midrule
JTT{\footnotesize~\citep{liu2021just}}              & \xmark  & \xmark &  ---         & 62.5\%$^\dagger$      &   ---     & 40.6\%$^\dagger$ & 81.3\%{\phantom{$^\dagger$}}   & 64.4\%{\phantom{$^\dagger$}}    \\\midrule
Group DRO
& \xmark  & \xmark \\
(with splits identified by \texttt{ls})\hspace{-20mm} & & & 91.2\%{\phantom{$^\dagger$}} & \textbf{86.1\%}{\phantom{$^\dagger$}}  & 87.2\%{\phantom{$^\dagger$}} & \textbf{83.3\%}{\phantom{$^\dagger$}} & 78.7\%{\phantom{$^\dagger$}}   & \textbf{72.1\%}{\phantom{$^\dagger$}}\\
\midrule\bottomrule
\end{tabular}
\end{table*}

%% file: sections/conclusion.tex
\section{Discussion}\label{sec:discussion}
Section~\ref{sec:applications} shows that \texttt{ls} identifies non-generalizable splits that correlate with human-identified biases. However, we must keep in mind that bias is a \emph{human-defined} notion. Given the set of input-label pairs, \texttt{ls} provides a tool for understanding potential biases, not a fairness guarantee. If the support of the given dataset doesn't cover the minority groups, \texttt{ls} will fail. For example, consider a dataset with only samoyeds in grass and polar bears in snow (no samoyeds in snow or polar bears in grass). \texttt{ls} will not be able to detect the background bias in this case.

We also note that poor generalization can result from label noise. Since the Splitter makes its decision based on the \emph{input-label} pair, \texttt{ls} can achieve high generalization gap by allocating all clean examples to the training split and all mislabeled examples to the testing split. Here we can think of \texttt{ls} as a label noise detector
(see Appendix~\ref{app:noisy} for more analysis). 
Blindly maximizing the worst-split performance in this situation will enforce the model to memorize the noise.

Finally, we present \texttt{ls} in the context of classification tasks. For regression tasks, we can sort examples in the testing split based on their mean square error and use this signal to guide the Splitter (Eq~\ref{eq:gap}). In real-world applications, biases can also come from many independent sources (e.g., gender and race).
Identifying multiple \emph{diverse} splits is another interesting future work.

\section{Conclusion}\label{sec:conclusion}
We present Learning to Split (\texttt{ls}), an algorithm that learns to split the data so that predictors trained on the training split cannot generalize to the testing split. Our algorithm only requires access to the set of input-label pairs and is applicable to general datasets. Experiments across multiple modalities confirm that \texttt{ls} identifies challenging splits that correlate with human-identified biases.
Compared to previous state-of-the-art, learning with \texttt{ls}-identified splits significantly improves robustness.

%% file: appendix/experiments.tex
\newpage
\appendix
\section{Datasets and model architectures}\label{app:data}
\subsection{Beer Review}
\paragraph{Data}
We use the BeerAdvocate review dataset~\citep{mcauley2012learning} and consider three \emph{binary} aspect-level sentiment classification tasks: \textsc{look}, \textsc{aroma} and \textsc{palate}. This dataset was originally downloaded from \url{https://snap.stanford.edu/data/web-BeerAdvocate.html}. 

\citet{lei2016rationalizing} points out that there exist strong correlations between the ratings of different aspects. In fact, the average correlation between two different aspects is 0.635. These correlations constitute as a source of biases when we apply predictors to examples with conflicting aspect ratings (e.g. beers that looks great but smells terrible).

We randomly sample 2500 positive examples and 2500 negative examples for each task. We apply \texttt{ls} to identify non-generalizable splits across these 5000 examples. The average word count per review is 128.5.

\paragraph{Representation backbone}
Following previous work~\citep{yujia2021predict}, we use a simple text CNN for this dataset. Specifically, each input review is encoded by pre-trained FastText embeddings~\cite{mikolov2018advances}. We employ 1D convolutions (with filter sizes $3, 4, 5$) to extract the features~\cite{kim-2014-convolutional}. We use 50 filters for each filter size. We apply max pooling to obtain the final representation ($\in\mathbb{R}^{150}$) for the input.

\paragraph{Predictor}
The Predictor applies a multi-layer perceopton on top of the previous input representation to predict the binary label. We consider a simple MLP with one hidden layer (150 hidden units). We apply ReLU activations and dropout (with rate 0.1) to the hidden units.

\paragraph{Splitter}
The Splitter concatenates the CNN representation with the binary input label. Similar to the Predictor, we use a MLP with one hidden layer (150 ReLU units, dropout 0.1) to predict the splitting decision $\mathbb{P}(z_i \mid x_i, y_i)$. We note that the representation backbones of the Splitter and the Predictor are \emph{not shared} during training.

\subsection{Tox21}
\paragraph{Data}
The dataset contains 12,707 chemical compounds. Each example is annotated with two types of properties: Nuclear Receptor Signaling Panel (AR, AhR, AR-LBD, ER, ER-LBD, aromatase, PPAR-gamma) and Stress Response Panel (ARE, ATAD5, HSE, MMP, p53).
The dataset is publicly available at \url{http://bioinf.jku.at/research/DeepTox/tox21.html}.

\paragraph{Representation backbone}
Following \cite{10.3389/fenvs.2015.00080}, we encode each input molecule by its dense features (such as molecular weight, solubility or surface area) and sparse features (chemical substructures). There are 801 dense features and 272,776 sparse features. We concatenate these features and standardize them by removing the mean and scaling to unit variance.

\paragraph{Predictor}
The Predictor is a multi-layer perceptron with three hidden layers (each with 1024 units). We apply ReLU activations and dropout (with rate 0.3) to the hidden units.

\paragraph{Splitter}
The Splitter concatenates the molecule features with the binary input label. Similar to the Predictor, we use a multi-layer perceptron with three hidden layers (each with 1024 units). We apply ReLU activations and dropout (with rate 0.3) to the hidden units.

\subsection{Waterbird}
\paragraph{Data}
This dataset is constructed from the CUB bird dataset~\citep{WelinderEtal2010} and the Places dataset~\citep{zhou2017places}.
\citet{sagawa2019distributionally} use the provided pixel-level segmentation information to crop each bird out from the its original background in CUB. The resulting birds are then placed onto different backgrounds obtained from Places. They consider two types of backgrounds: water (ocean or natural lake) and land (bamboo forest or broadleaf forest). There are 4795/1199/5794 examples in the training/validation/testing set. This dataset is publicly available at \url{https://nlp.stanford.edu/data/dro/waterbird_complete95_forest2water2.tar.gz}

By construction, 95\% of all waterbirds in the training set have water backgrounds. Similarly, 95\% of all landbirds in the training set have land backgrounds. As a result, predictors trained on this training data will overfit to the spurious background information when making their predictions. In the validation and testing sets, \citet{sagawa2019distributionally} place landbirds and waterbirds equally to land and water backgrounds.

For identifying non-generalizable splits, we apply \texttt{ls} on the training set and the validation set. For automatic de-biasing,
we report the average accuracy and worst-group accuracy on the official test set. To compute the worst-group accuracy, we use the background attribute to partition the test set into four groups: 
\texttt{waterbirds} with water backgrounds,
\texttt{waterbirds} with land backgrounds,
\texttt{landbirds} with water backgrounds,
\texttt{landbirds} with land backgrounds.

\paragraph{Representation backbone}
Following previous work~\citep{sagawa2019distributionally, liu2021just},
we fine-tune torchvision's \texttt{resnet-50}, pretrained on ImageNet~\citep{imagenet_cvpr09}, to represent each input image. This results into a 2048 dimensional feature vector for each image.

\paragraph{Predictor}
The Predictor takes the \texttt{resnet} representation and applies a linear layer (2048 by 2) followed by Softmax to predict the label ($\{\texttt{waterbirds},\, \texttt{landbirds}\}$) of each image.
Note that we reset the Predictor's parameter to the pre-trained \texttt{resnet-50} at the beginning of each outer-loop iteration.

\paragraph{Splitter}
The Splitter first concatenates the \texttt{resnet} representation with the binary image label. It then applies a linear layer with Softmax to predict the splitting decision $\mathbb{P}(z_i \mid x_i, y_i)$. The \texttt{resnet} encoders for the Splitter and the Predictor are not shared during training. 

\subsection{CelebA}
\paragraph{Data}
CelebA~\citep{liu2015deep} is a large-scale face attributes dataset, where each image is annotated with 40 binary attributes. Following previous work~\citep{sagawa2019distributionally, liu2021just}, we consider our task as predicting the blond hair attribute ($\in\{\texttt{blond\_hair}, \texttt{no\_blond\_hair}\}$). The CelebA dataset is available for non-commercial research purposes only. It is publicly available at \url{https://mmlab.ie.cuhk.edu.hk/projects/CelebA.html}.

While there are lots of annotated examples in the training set (162,770), the task is challenging due to the spurious correlation between the target blond hair attribute and the gender attribute ($\in\{\texttt{male}, \texttt{female}\}$). Specifically, only 0.85\% of the training data are blond-haired males. As a result, predictors learn to utilize \texttt{male} as a predictive feature for \texttt{no\_blond\_hair} when we directly minimizing their empirical risk.

For identifying non-generalizable splits, we apply \texttt{ls} on the official training set and validation set. For automatic de-biasing, we report the average and worst-group performance on the official test set. To compute the worst-group accuracy, we use the gender attribute to partition the test set into four groups: 
\texttt{blond\_hair} with male,
\texttt{blond\_hair} with female,
\texttt{no\_blond\_hair} with male,
\texttt{no\_blond\_hair} with female.

\paragraph{Representation backbone}
Following previous work~\citep{sagawa2019distributionally, liu2021just},
we fine-tune torchvision's \texttt{resnet-50}, pretrained on ImageNet~\citep{imagenet_cvpr09}, to represent each input image. This results into a 2048 dimensional feature vector for each image.

\paragraph{Predictor}
The Predictor takes the \texttt{resnet} representation and applies a linear layer (2048 by 2) followed by Softmax to predict the label ($\{\texttt{blond\_hair},\, \texttt{no\_blond\_hair}\}$) of each image.
Note that we reset the Predictor's parameter to the pre-trained \texttt{resnet-50} at the beginning of each outer-loop iteration.

\paragraph{Splitter}
The Splitter concatenates the \texttt{resnet} representation with the binary image label. It then applies a linear layer with Softmax to predict the splitting decision $\mathbb{P}(z_i \mid x_i, y_i)$. The \texttt{resnet} encoders for the Splitter and the Predictor are not shared during training.

\subsection{MNLI}
\paragraph{Data}
The MultiNLI corpus contains 433k sentence pairs~\citep{N18-1101}. Given a sentence pair, the task is to predict the entailment relationship (\texttt{entailment}, \texttt{contradiction}, \texttt{neutral}) between the two sentences. The original corpus splits allocate most examples to the training set, with another 5\% for validation and the last 5\% for testing. In order to accurately measure the performance on rare groups, \citet{sagawa2019distributionally} combine the training and validation set and randomly shuffle them into a 50/20/30 training/validation/testing split. 
The dataset and splits are publicly available at \url{https://github.com/kohpangwei/group_DRO}.

Previous work~\citep{gururangan2018annotation, mccoy-etal-2019-right} have shown that this crowd-sourced dataset has significant annotation artifacts: negation words (nobody, no, never and nothing) often appears in \texttt{contradiction} examples; sentence pairs with high lexical overlap are likely to be \texttt{entailment}. As a result, predictors may over-fit to these spurious shortcuts during training.

For identifying non-generalizable splits, we apply \texttt{ls} on the training set and validation set. For automatic de-biasing, we report the average and worst-group performance on the testing set. To compute the worst-group accuracy, we partition the test set based on whether the input example contains negation words or not:
\texttt{entailment} with negation words,
\texttt{entailment} without negation words,
\texttt{contradiction} with negation words,
\texttt{contradiction} without negation words,
\texttt{neutral} with negation words,
\texttt{neutral} without negation words.

\paragraph{Representation backbone}
Following previous work~\citep{sagawa2019distributionally, liu2021just},
we fine-tune Hugging Face's \texttt{bert-base-uncased} model, starting with pre-trained weights~\citep{47751}.

\paragraph{Predictor}
The Predictor takes the representation of the \texttt{[CLS]} token (at the final layer of \texttt{bert-base-uncased}) and applies a linear layer with Softmax activations to predict the final label (\texttt{entailment}, \texttt{contradictions}, \texttt{neutral}).
Note that we reset the Predictor's parameter to the pre-trained \texttt{bert-base-uncased} at the beginning of each outer-loop iteration.

\paragraph{Splitter}
The Splitter concatenates the representation of the \texttt{[CLS]} token with the \emph{one-hot} label embedding ($\in\{0,1\}^{3}$). It then applies a linear layer with Softmax activations to predict the splitting decision $\mathbb{P}(z_i \mid x_i, y_i)$. The \texttt{bert-base-uncased} encoders for the Splitter and Predictor are not shared during training. 

%% file: appendix/details.tex
\section{Implementation details}
\label{app:details}
\subsection{Identifying non-generalizable splits using \texttt{ls}}
\label{app:details_ls}
\paragraph{Optimization}
For Beer Review, Tox21, Waterbirds and CelebA, we update the Splitter and Predictor with the Adam optimizer~\citep{kingma2014adam}. In Beer Review and Tox21, the learning rate is set to $10^{-3}$ with no weight decay (as we already have dropout in the MLP to prevent over-fitting). We use a batch size of 200.
In Waterbirds and CelebA, since we start with pre-trained weights, we adopt a smaller learning rate $10^{-4}$~\citep{sagawa2019distributionally} and set weight decay to $10^{-3}$. We use a batch size of 100.
For MNLI, we use the default setting for fine-tuning BERT:
a fixed linearly-decaying learning rate starting at 0.0002, AdamW optimizer~\citep{loshchilov2017decoupled}, dropout, and no weight decay. We use a batch size of 100.

\paragraph{Stopping criteria}
For the Predictor's training, we held out a random 1/3 subset of $\Dtrain$ for validation. We train the Predictor on the rest of $\Dtrain$ and apply early-stopping when the validation accuracy stops improving in the past 5 epochs.
For the Splitter's training, we compare the average loss $\mathcal{L}^\text{total}$ of the current epoch and the average loss across the past 5 epochs. We stop training if the improvement is less than $10^{-3}$.


\subsection{Automatic de-biasing}
\paragraph{Method details}
We use the Splitter learned by \texttt{ls} to create groups that are informative of the biases. Specifically, for each example $(x_i, y_i)$, we first sample its splitting decision from the Splitter $\hat{z}_i \sim \mathbb{P}(z_i \mid x_i, y_i)$. As we have seen in Figure~\ref{fig:analysis}, these splitting decisions reveal human-identified biases. Similar to the typical group DRO setup~\citep{sagawa2019distributionally}, we use these information together with the target labels to partition the training and validation data into different groups. For example in Waterbirds, we have four groups: $\{y=\texttt{waterbirds}, z=0\}, \{y=\texttt{waterbirds}, z=1\}, \{y=\texttt{landbirds}, z=0\}, \{y=\texttt{landbirds}, z=1\}$. 
We minimize the worst-group loss during training and measure the worst-group accuracy on the validation data for model selection.
Specifically, we stop training if the validation metric hasn't improved in the past 10 epochs.

\paragraph{Optimization}
Modern neural networks are usually highly over-parameterized. As a result, they can easily memorize the training data and over-fit the majority groups even when we minimize the worst-group loss during training. Following \citet{sagawa2019distributionally}, we apply strong regularization to combat memorization and over-fitting. We grid-search over the weight decay parameter ($10^0, 10^{-1}, 10^{-2}, 10^{-3}, 0$).

\subsection{Computing resources}\label{app:resource}
We conducted all the experiments on our internal clusters with NVIDIA A100, NVIDIA RTX A6000, and NVIDIA Tesla V100.

%% file: appendix/noisy.tex
\section{Application: label noise detection}
\label{app:noisy}
In the presence of label noise, \texttt{ls} can reach high generalization gap by allocating all clean examples to the training split and all mislabeled examples to the testing split. Here we verify the effectiveness of \texttt{ls} as a label noise detector.

\paragraph{Data}
We consider the standard digit classification dataset MNIST as our test bed~\citep{lecun-mnisthandwrittendigit-2010}. The dataset is freely available at ~\url{http://yann.lecun.com/exdb/mnist/}.

We use the official training data (10 classes and 60,000 examples in total) and inject random label noise.
Specifically, for a given noise ratio $\eta$, we sample a noisy label $\tilde{y}$ of each image based on its original clean label $y$:
\[
\mathbb{P}(\tilde{y}) = 
\left\{\begin{array}{lr}
        1-\eta & \text{for } \tilde{y}=y,\\
        \eta / 9 & \text{for } \tilde{y}\neq y.
        \end{array}\right.
\]
We apply \texttt{ls} to identify spurious splits from this noisy data collection for various $\eta$.
In practice, we cannot assume access to the noise ratio.
Therefore we keep $\delta=0.75$ (Eq~\ref{eq:reg}) as in our other experiments.

\paragraph{Representation backbone}
We follow the architecture from PyTorch's MNIST example\footnote{https://github.com/pytorch/examples/blob/master/mnist/main.py}.
Each input image is passed to a CNN with 2 convolution layers followed by max pooling ($2\times 2$).
The first convolution layer has 32 filters, and the second convolution layer has 64 filters.
Filter sizes are set to $3\times 3$ in both layers.

\paragraph{Predictor}
The Predictor is a multi-layer perceptron with two hidden layers (each with 100 units). We apply ReLU activations and dropout (with rate 0.25) to the hidden units.

\paragraph{Splitter}
The Splitter concatenates the CNN features with the one-hot input label. Similar to the Predictor, we use a multi-layer perceptron with two hidden layers (each with 100 units). We apply ReLU activations and dropout (with rate 0.25) to the hidden units.

\paragraph{Optimization}
The optimization strategy is the same as the one for Tox 21 (Section~\ref{app:details_ls}).

\paragraph{Evaluation metrics}
Given the spurious splits produced by \texttt{ls}, we can evaluate its precision and recall of identifying the polluted annotations:\looseness=-1
\begin{itemize}
    \item Precision = \#polluted annotations in the testing split / \#annotations in the testing split;
    \item Recall = \#polluted annotations in the testing split / \#polluted annotations.
\end{itemize}
We note that \texttt{ls} controls the train-test split ratio through the regularity constraint ($\Omega_1$ in Eq\ref{eq:reg}). For ease of optimization, this constraint is implemented as a \emph{soft} regularizer. As a result, the train-test split ratio needs to compete with other objectives (such as maximizing the generalization gap), and the resulting split ratio can be different for different noise ratios.
To better understand the precision and recall, given the train-test split ratio produced by \texttt{ls}, we define an oracle which allocates as many polluted annotations as possible into the testing split:
 
\begin{itemize}
    \item if \#polluted annotations $\leq$ \#annotations in the testing split:
    \begin{itemize}
        \item Oracle precision = \#polluted annotations / \#annotations in the testing split;
        \item Oracle recall = 100\%;
    \end{itemize}
    \item else:
    \begin{itemize}
        \item Oracle precision = 100\%;
        \item Oracle recall = \#annotations in the testing split / \#polluted annotations.
    \end{itemize}
\end{itemize}

\begin{figure}[t]
     \centering
     \includegraphics[width=\linewidth]{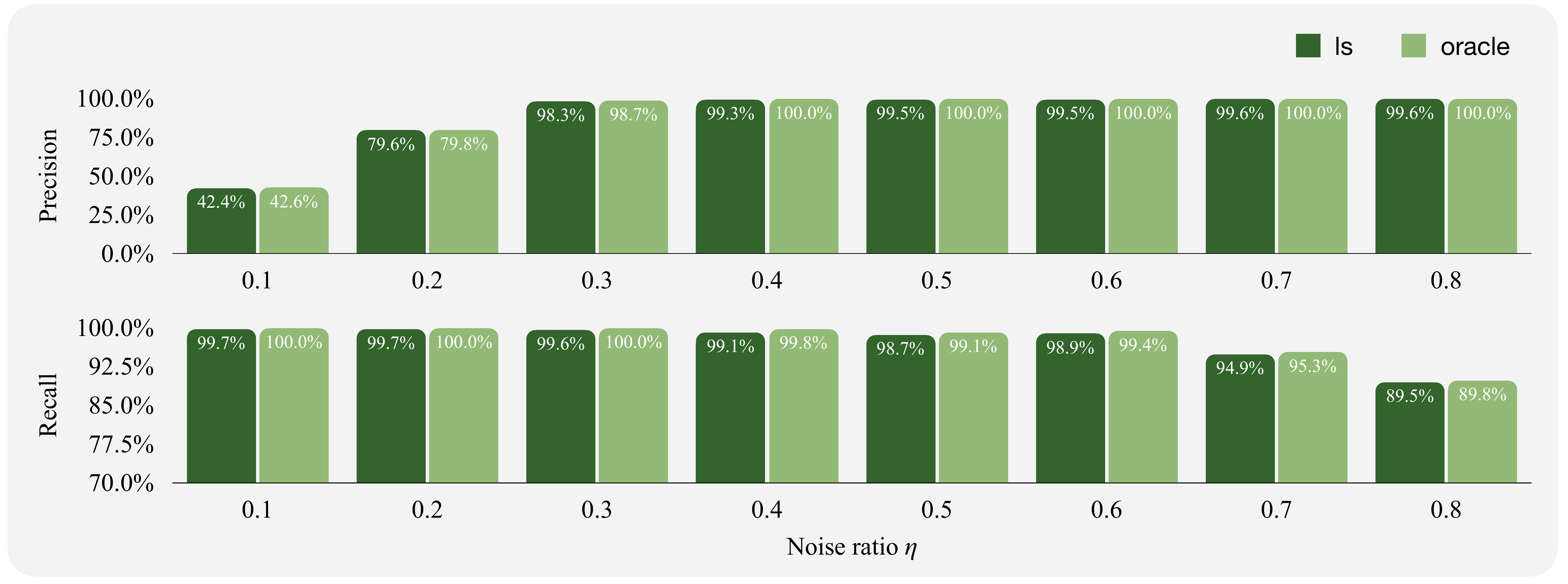}
     \caption{Noise}
     \label{fig:noise}
\end{figure}

\paragraph{Results}
We consider different noise ratios and present our results in Figure~\ref{fig:noise}. We observe that $\texttt{ls}$ consistently approaches to the oracle performance.
When the noise ratio $\eta$ is small ($0.1, 0.2$), \texttt{ls} allocates most of the polluted annotations (>99\%) into the testing split. However, to meet the train-test ratio constraint, it also has to include some of the clean annotations into the testing split (non-perfect precision). When the noise ratio $\eta$ is large ($0.7, 0.8$), the testing split consists of mostly polluted annotations (>99\%). The recall is not perfect because allocating all polluted annotations to the testing split will violate the regularity constraint.

%% file: appendix/full.tex
\section{Additional results}
\label{app:full}

\begin{figure}[h]
     \centering
     \includegraphics[width=0.97\linewidth]{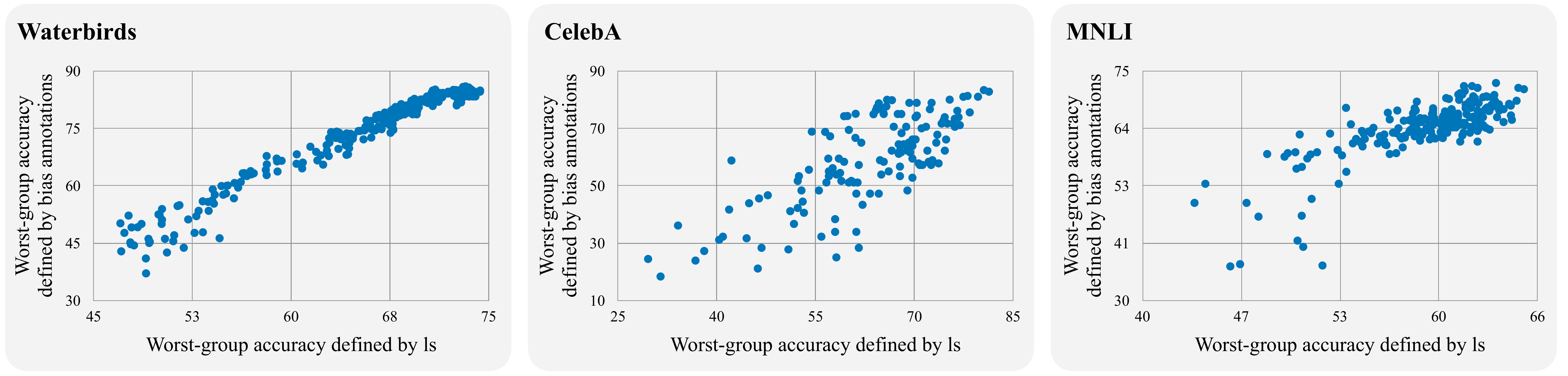}
     \caption{The spurious splits identified by \texttt{ls} provide a surrogate metric for model selection when biases are not known a priori.
     }
     \label{fig:surrogate}
\end{figure}